\documentclass[10pt,conference]{IEEEtran}

\usepackage[utf8]{inputenc} 
\usepackage[T1]{fontenc}    
\usepackage{hyperref}       
\usepackage{url}            
\usepackage{booktabs}       
\usepackage{amsfonts}       
\usepackage{nicefrac}       
\usepackage{microtype}      
\usepackage{amsmath}
\usepackage{amssymb}
\usepackage{amsthm}
\usepackage{xcolor}
\usepackage{tikz}
\usepackage{graphicx}
\usepackage{cite}
\usepackage{algorithm}
\usepackage[noend]{algpseudocode}

\floatname{algorithm}{Procedure}

\usetikzlibrary{arrows,shapes,chains,matrix,positioning,scopes,automata,backgrounds,calc,shadows.blur,through}

\begin{document}

\title{Neurogenesis Deep Learning \\ {\large Extending deep networks to accommodate new classes}}
\author{ \IEEEauthorblockN{Timothy J.~Draelos\IEEEauthorrefmark{1}, 
Nadine E.~Miner\IEEEauthorrefmark{1}, 
Christopher C.~Lamb\IEEEauthorrefmark{1}, 
Jonathan A.~Cox\IEEEauthorrefmark{1}\IEEEauthorrefmark{2}, \\
Craig M.~Vineyard\IEEEauthorrefmark{1}, 
Kristofor D.~Carlson\IEEEauthorrefmark{1}, 
William M.~Severa\IEEEauthorrefmark{1}, 
Conrad D.~James\IEEEauthorrefmark{1}, 
and James B.~Aimone\IEEEauthorrefmark{1}}
\IEEEauthorblockA{\IEEEauthorrefmark{1}Sandia National Laboratories Albuquerque, NM, USA \\  \{{\ttfamily tjdrael}, {\ttfamily nrminer}, {\ttfamily cclamb}, {\ttfamily cmviney}, {\ttfamily kdcarls}, {\ttfamily wmsever}, {\ttfamily cdjame}, {\ttfamily jbaimon}\}{\ttfamily@sandia.gov }} \IEEEauthorblockA{ \IEEEauthorrefmark{2} Present Address: Qualcomm Corporation, San Diego, CA, USA {\ttfamily joncox@alum.mit.edu}}}
\maketitle
\begin{abstract}
Neural machine learning methods, such as deep neural networks (DNN), have achieved remarkable success in a number of complex data processing tasks. These methods have arguably had their strongest impact on tasks such as image and audio processing – data processing domains in which humans have long held clear advantages over conventional algorithms. In contrast to biological neural systems, which are capable of learning continuously, deep artificial networks have a limited ability for incorporating new information in an already trained network. As a result, methods for continuous learning are potentially highly impactful in enabling the application of deep networks to dynamic data sets. Here, inspired by the process of adult neurogenesis in the hippocampus, we explore the potential for adding new neurons to deep layers of artificial neural networks in order to facilitate their acquisition of novel information while preserving previously trained data representations. Our results on the MNIST handwritten digit dataset and the NIST SD $\mathbf{19}$ dataset, which includes lower and upper case letters and digits, demonstrate that neurogenesis is well suited for addressing the stability-plasticity dilemma that has long challenged adaptive machine learning algorithms.

\end{abstract}

\section{Introduction}
Machine learning methods are powerful techniques for statistically extracting useful information from “big data” throughout modern society. In particular, deep learning (DL) and other deep neural network (DNN) methods have proven successful in part due to their ability to utilize large volumes of unlabeled data to progressively form sophisticated hierarchical abstractions of information~\cite{lecun2015deep,schmidhuber2015deep}. While DL's training and processing mechanisms are quite distinct from biological neural learning and behavior, the algorithmic structure is somewhat analogous to the visual processing stream in mammals in which progressively deeper layers of the cortex appear to form more abstracted representations of raw sensory information acquired by the retina~\cite{van1992information}.

DNNs are typically trained once, either with a large amount of labelled data or with a large amount of unlabeled data followed by a smaller amount of labeled data used to “fine-tune” the network for some particular function, such as handwritten digit classification. This training paradigm is often very expensive, requiring several days on large computing clusters~\cite{le2013building}, so ideally a fully trained network will continue to prove useful for a long duration even if the application domain changes. DNNs have found some successes in transfer learning, due to their general-purpose feature detectors at shallow layers of a network~\cite{cirecsan2012transfer, yosinski2014transferable}, but our focus is on situations where that is not the case. DNNs' features are known to get more specialized at deeper layers of a network and therefore presumably less robust to new classes of data. In this work, we focus on inputs that a trained network finds difficult to represent. In this regard, we are addressing the problem of continuous learning (CL). In reality, DNNs may not be robust to concept drift, where the data being processed changes gradually over time (e.g.,~a movie viewer's preferred genres as they age), nor transfer learning, where a trained model is repurposed to operate in a different domain. Unlike the developing visual cortex, which is exposed to varying inputs over many years, the data used to train DNNs is typically limited in scope, thereby diminishing the applicability of networks to encode information statistically distinct from the training set. The impact of such training data limitations is a relatively minor concern in cases where the application domain does not change (or changes very slowly). However, in domains where the sampled data is unpredictable or changes quickly, such as what is seen by a cell phone camera, the value of a static deep network may be quite limited. 

One mechanism the brain has maintained in selective regions such as the hippocampus is the permissive birth of new neurons throughout one's lifetime, a process known as adult neurogenesis~\cite{aimone2014regulation}. While the specific function of neurogenesis in memory is still debated, it clearly provides the hippocampus with a unique form of plasticity that is not present in other regions less exposed to concept drift. The process of biological neurogenesis is complex, but two key observations are that new neurons are preferentially recruited in response to behavioral novelty and that new neurons gradually learn to encode information (e.g.,~they are not born with pre-programmed representations, rather they learn to integrate over inputs during their development)~\cite{aimone2011resolving}.

We consider the benefits of neurogenesis on DL by exploring whether ``new'' artificial neurons can facilitate the learning of novel information in deep networks while preserving previously trained information. To accomplish this, we consider a specific illustrative example with the MNIST handwritten digit dataset~\cite{lecun1998gradient} and the larger NIST SD $19$ dataset~\cite{grother1995nist} that includes handwritten digits as well as upper and lower case letters. An autoencoder (AE) is initially trained with a subset of a dataset's classes and continuous adaptation occurs by learning each remaining class. Our results demonstrate that neurogenesis with hippocampus-inspired ``intrinsic replay'' (IR) enables the learning of new classes with minimal impairment of original representations, which is a challenge for conventional approaches that continue to train an existing network on novel data without structural changes.
\subsection{Related Work}
In the field of machine learning, transfer learning addresses the problem of utilizing an existing trained system on a new dataset containing objects of a different kind. Over the past few years, researchers have examined different ways of transferring classification capability from established networks to new tasks. Recent approaches have taken a horizontal approach, by transferring layers, rather than a more finely grained vertically oriented approach of dynamically creating or eliminating individual nodes in a layer. Neurogenesis has been proposed to enable the acquisition of novel information while minimizing the potential disruption of previously stored information~\cite{aimone2011resolving}. Indeed, neurogenesis and similar processes have been shown to have this benefit in a number of studies using shallow neural networks~\cite{appleby2011role, appleby2009additive, carpenter1988art, chambers2007network, chambers2004simulated, crick2006apoptosis,wiskott2006functional,aimone2011modeling}, although these studies have typically focused on more conventional transfer learning, as opposed to the continuous adaptation to learning considered here.

An adaptive DNN architecture by Calandra, et al, shows how DL can be applied to data unseen by a trained network~\cite{calandra2012learning}. Their approach hinges on incrementally re-training deep belief networks (DBNs) whenever concept drift emerges in a monitored stream of data and operates within constant memory bounds. They utilize the generative capability of DBNs to provide training samples of previously learned classes. Class conditional sampling from trained networks has biological inspiration~\cite{carr2011hippocampal,louie2001temporally,stickgold2005sleep,felleman1991distributed} as well as historical and artificial neural network implementations~\cite{hinton1995wake, salakhutdinov2009learning, gregor2015draw, rudy2014generative}.

Yosinski evaluated transfer capability via high-level layer reuse in specific DNNs~\cite{yosinski2014transferable}. Transferring learning in this way increased recipient network performance, though the closer the target task was to the base task, the better the transfer. Transferring more specific layers could actually cause performance degradation however. Likewise, Kandaswamy, et al., used layer transfer as a means to transfer capability in Convolutional Neural Networks and Stacked Denoising AEs~\cite{kandaswamy2014improving}. Transferring capability in this way resulted in a reduction in overall computation time and lower classification errors. These papers use fixed-sized DNNs, except for additional output nodes for new classes, and demonstrate that features in early layers are more general than features in later layers and thus, more transferable to new classes.

\section{The Neurogenesis Deep Learning Algorithm}
\begin{figure}
\begin{center}
\includegraphics[width=3.5in]{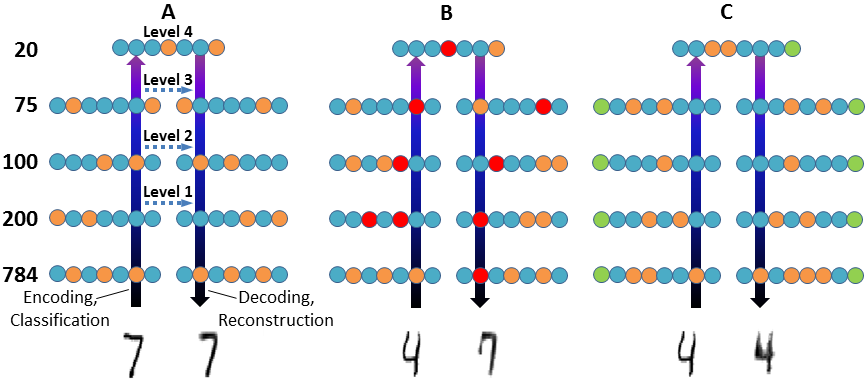}
\caption{\label{newNodesFigure} Illustration of NDL processing MNIST digits (orange/red circles indicate accuracte/inaccurate feature representations of the input; green indicate new nodes added via neurogenesis). (A) AE can faithfully reconstruct originally trained digit (`$7$'), but (B) fails at reconstructing novel digit (`$4$'). (C) New nodes added to all levels enables AE to reconstruct `$4$'. Level $1$--$4$ arrows show how inputs can be reconstructed at various depths.}
\end{center}
\end{figure}

Neurogenesis in the brain provides a motivation for creating DNNs that adapt to changing environments. Here, we introduce the concept of neurogenesis deep learning (NDL), a process of incorporating new nodes in any level of an existing DNN (Figure~\ref{newNodesFigure}) to enable the network to adapt as the environment changes. We consider the specific case of adding new nodes to pre-train a stacked deep AE, although the approach should extend to other types of DNNs as well. An AE is a type of neural network designed to encode data such that they can be decoded to produce reconstructions with minimal error. The goal of many DNN algorithms is to learn filters or feature detectors (i.e.,~weights) where the complexity or specialization of the features increases at deeper network layers. Although successive layers of these feature detectors could require an exponential expansion of nodes to guarantee that all information is preserved as it progresses into more sophisticated representations (``lossless encoding''), in practice, deep AEs typically use a much more manageable number of features by using the training process to select those features that best describe the training data. However, there is no guarantee that the representations of deeper layers will be sufficient to losslessly encode novel information that is not representative of the original training set. It is in this latter case that we believe NDL to be most useful, as we have previously suggested that biological neurogenesis addresses a similar coding challenge in the brain~\cite{aimone2011resolving}.

The first step of the NDL algorithm occurs when a set of new data points fail to be appropriately reconstructed by the trained network. A reconstruction error (RE) is computed at each level of a stacked AE (pair of encode/decode layers) to determine when a level's representational capacity is considered insufficient for a given application. An AE parameterized with weights, $W$, biases, $b$, and activation function, $s$, is described from input, $x$, to output as $N$ encode layers followed by $N$ decode layers.
\begin{align}
\scriptstyle \text{Encoder: }  f_{\theta_N} \circ f_{\theta_{N-1}} \cdots f_{\theta_2} \circ f_{\theta_1}(x) \text{ where }& \scriptstyle y = f_\theta(x) = s(Wx + b)\\
\scriptsize \text{Decoder: } \scriptstyle g_{\theta'_N} \circ g_{\theta'_{N-1}} \cdots g_{\theta'_2} \circ g_{\theta'_1}(y) \text{ where }& \scriptstyle g_{\theta'}(y) = s(W'y + b') \label{eq2}
\end{align}
Global RE is computed at level $L$ of an AE by encoding an input through $L$ encode layers, then propagating through the corresponding $L$ decode layers to the output. 
\begin{align}
\scriptstyle RE_{Global, L} (x) = (x - g_{\theta'_N} \circ \cdots g_{\theta'_{N-L}} \circ f_{\theta_L} \circ \cdots f_{\theta_1}(x))^2
\end{align}
When a data sample's RE is too high, the assumption is that the AE level under examination does not contain a rich enough set of nodes (or features as determined by each node's weights) to accurately reconstruct the sample. Therefore, it stands to reason that a sufficiently high RE warrants the addition of a new feature detector (node).

The second step of the NDL algorithm is adding and training new nodes, which occurs when a critical number of input samples (outliers) fail to achieve adequate representation at some level of the network. A new node is also added if the previous level added one or more nodes. This process does not require labels, relying entirely on the quality of a sample's representation computed from its reconstruction. If the RE is too high (greater than a user-specified threshold determined from the statistics of reconstructing previously seen data classes), then nodes are added at that level up to a user-specified maximum number of new nodes. The new nodes are trained using all nodes in the level for reconstruction on all outliers. In other words, during training of the new nodes, the reconstructions, errors, gradients, and weight updates are calculated as a function of an AE that uses the entire set of nodes in the current level within a single hidden layer AE (SHL-AE). In order to not disturb the existing feature detectors, only the encoder weights connected to the new nodes are updated in the level under consideration. Decoder weights connected to existing feature detectors (nodes) are allowed to change slightly at the learning rate divided by $100$.

\begin{figure}
\begin{center}
\includegraphics[width=2.5in]{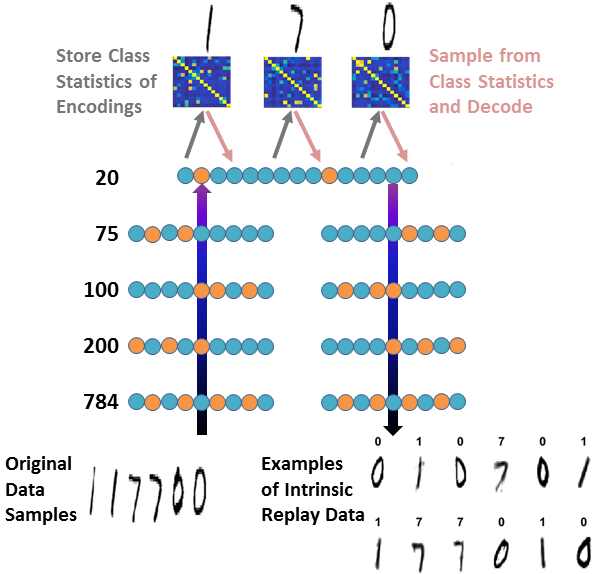}
\caption{\label{irFigure} Illustration of the intrinsic replay process used in NDL. Original data presented to a trained network results in high-level representations in the ``top-most'' layer of the encoder. The average entries and the Cholesky decomposition of the covariance matrix of this hidden layer are stored for each class (e.g., `$1$'s, `$7$'s, and `$0$'s). When ``replayed'' values are desired for a given class, samples are drawn randomly from a normal distribution defined by the class's stored statistics. Then, using the AE's reconstruction pathway, new digits of the stored class are approximated.}
\end{center}
\end{figure}

The final step of the NDL algorithm is intended to stabilize the network's previous representations in the presence of newly added nodes. It involves training all nodes in a level with new data and replayed samples from previously seen classes on which the network has been trained. Samples from old classes, where original data no longer exists, are created using the encoding and reconstruction capability of the current network in a process we call ``intrinsic replay'' (IR) (Figure~\ref{irFigure}).

This IR process is analogous to observed dynamics within the brain's hippocampus during memory consolidation~\cite{carr2011hippocampal}. It appears that neural regions such as the hippocampus ``replay'' neuronal sequences originally experienced during learned behaviors or explorations in an effort to strengthen and stabilize newly acquired information alongside previously encoded information. Our method involves storing class-conditional statistics (mean and Cholesky factorization of the covariance) of the top layer of the encoding network, $E$.
\begin{align}
\mu_E = \text{Mean}(E), Ch_E = \text{Chol}(\text{Cov}(E))
\end{align}
The Cholesky decomposition requires $n^3/6$ operations~\cite{krishnamoorthy2013matrix}, where $n$ is the dimension of $E$, and is performed once for each class on a trained network. High-level representations are retrieved through sampling from a Normal distribution described by these statistics and, leveraging the decoding network, new data points from previously trained classes are reconstructed.
\begin{align}
\text{IR Images } = \text{Decode}(\mu_E + N(0,1)*Ch_E) \label{eq5}
\end{align}
Training samples from previously seen data classes, where original data no longer exists, are generated using (\ref{eq5}), which involves a single feed-forward pass through the Decoder (\ref{eq2}).
\begin{algorithm} \footnotesize
\caption{Neurogenesis Deep Learning (NDL) \label{algor}}
\begin{algorithmic}
\State \textbf{Input: } $2N$-layer $AE$ trained on data classes $D_1$--$D_{U-1}$, new class of data $D_U$, vector of per-level RE thresholds $Th$, vector of per-level maximum nodes allowed $MaxNodes$, maximum number of samples allowed to have $RE_{Global,L} > Th_L$, $MaxOutliers$, Learning Rate $LR$

\vspace*{.1in}
\State \textbf{Output: } Autoencoder $AE$ capable of representing classes $D_1$--$D_U$

\vspace*{.1in} 
\State // Create stabilization training data
\State $AE_{StableTrain} \leftarrow \{D_U | \text{IntrinsicReplay}(D_1\text{--}D_{U-1}) \}$

\vspace*{.1in} 
\State // Perform Neurogenesis
\For{$Level L \leftarrow 1$ to $N$}
\State $NewNodes \leftarrow 0$
\State $Outliers \leftarrow \{d \ni D_U | RE_{Global, L}(d) > Th_L\}$
\State $N_{Out} \leftarrow |Outliers|$

\vspace*{.1in} 
\State // Add new nodes to $AE_L$ and train
\While{$N_{Out} > MaxOutliers$ and $NewNodes < MaxNodes_L$}
\State $AE_L \leftarrow W_L, b_L ; W'_{N+1-L}, b'_{N+1-L}$ from $AE$
\State \textbf{Plasticity: } $Nodes_{New} = \text{\# of new nodes to add}$
\State Add $Nodes_{New}$ to $AE_L$ and Train on $Outliers$
\State Use $LR$ to update encoder weights connected to new nodes only
\State Use $LR/100$ to update decoder weights
\State  \parbox[t]{\dimexpr\linewidth-\algorithmicindent-\algorithmicindent}{\textbf{Stability: } Train $AE_L$ on $AE_{StableTrain}$ using $LR/100$ to update all weights}
\State $W_L, b_L; W'_{N+1-L}, b'_{N+1-L} \leftarrow AE_L$
\State $Outliers \leftarrow \{d \ni D_U | RE_{Global, L}(d) > Th_L \}$
\State $N_{Out} \leftarrow |Outliers|$
\State $NewNodes \leftarrow NewNodes + Nodes_{New}$
\EndWhile

\vspace*{.1in} 
\State \parbox[t]{\dimexpr\linewidth-\algorithmicindent}{// Add random weights from new nodes in level $L$ to existing nodes in level $L+1$ and train $AE_{L+1}$}
\If{$NewNodes > 0$ and $L < N$}
\State \textbf{Plasticity: } Train $AE_{L+1}$ on $D_U$
\State \textbf{Stability: } Train $AE_{L+1} $ on $AE_{StableTrain}$
\State $W_{L+1}, b_{L+1}; W'_{N-L}, b'_{N-L} \leftarrow AE_{L+1}$
\EndIf
\EndFor 
\end{algorithmic}
\end{algorithm}

\section{Experiments}
We evaluated NDL on two datasets, the MNIST~\cite{lecun1998gradient} and NIST SD $19$~\cite{grother1995nist} datasets. For the NIST dataset, we downsampled the original $128$x$128$ pixel images to be $28$x$28$ (the MNIST image size). However, we did not otherwise normalize the characters within classes, so the variation in scale and location within the $28$x$28$ frame is much greater than the MNIST data.

For the MNIST dataset, a deep AE was pre-trained in a stacked layered manner on a subset of the dataset classes, then training with and without NDL and with and without IR was conducted on new unseen data classes. The AE was initially trained with two digits ($1$, $7$) that are not statistically representative of the other digits (as shown in the results). Then, learning was incrementally performed with the remaining digits. We used an $8$-layer AE inspired by Hinton's network on MNIST~\cite{hinton2006reducing}, but reduced to $784$-$200$-$100$-$75$-$20$-$75$-$100$-$200$-$784$ since only a subset of digits ($1$, $7$) were used for initial training. For each experiment, all training samples in a class were presented at once. 

For the NIST SD $19$ dataset, the AE was trained on the digit classes alone ($0$--$9$), and then learning was performed incrementally on all letters (upper and lower case; A-Z, a-z). In order to evaluate the impact of NDL on the NIST dataset without the potentially complicating factor of IR, training data was used for replaying old classes. The initial AE used for the NIST SD $19$ dataset is also inspired by Hinton's MNIST network, where the only difference is the number of highest-level features. We used $50$ instead of $30$ high-level features since there is much more variation in scale and location in the NIST digits. The trained NIST SD $19$ `Digits' network is $784$-$1000$-$500$-$250$-$50$-$250$-$500$-$1000$-$784$.

\section{Results on MNIST}
\subsection{Trained networks have limited ability to represent novel information}
\begin{figure}
\begin{center}
\includegraphics[width=2.5in]{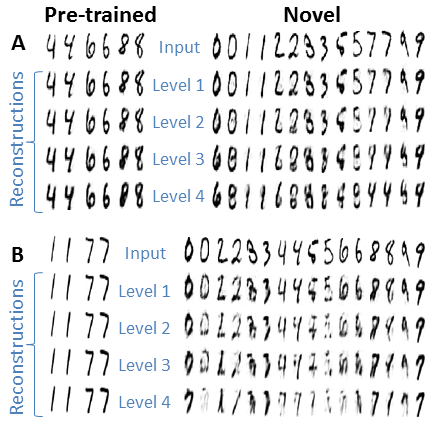}
\caption{Networks initially trained on (A) `$4$,' `$6$,' and `$8$'s and (B) `$1$,' and `$7$'s and not yet trained on any of the other MNIST digits reconstruct those novel digits using features biased by their original training data. \label{initialFigure}}
\end{center}
\end{figure}

To illustrate the process of NDL on MNIST data, we first trained a deep AE ($784$-$1000$-$500$-$250$-$30$-$250$-$500$-$1000$-$784$) to encode a subset of MNIST classes. Then, nodes were added via neurogenesis to the trained AE network as needed to encode each remaining digit. The initial DNN size for our illustrative example was determined as follows. In Calandra's work, a $784$-$600$-$500$-$400$-$10$ DBN classifier was trained initially on digits $4$, $6$, and $8$ and then presented with new digits for training together with samples of $4$, $6$, and $8$ generated from the DBN~\cite{calandra2012learning}. We examined two subsets of digits for initial training of our AE ($4$, $6$, and $8$, as in Calandra, et al.~\cite{calandra2012learning}, or $1$ and $7$). Figure~\ref{initialFigure}A illustrates that digits $4$, $6$, and $8$ appear to contain a more complete set of digit features as seen by the quality of the reconstructions compared to training only on $1$ and $7$ (Figure~\ref{initialFigure}B), although both limited training sets yield impaired reconstructions of novel (untrained) digits. We chose to focus initial training on digits $1$ and $7$, as these digits represent what may be the smallest set of features in any pair of digits. Then, continuous learning was simulated by progressively expanding the number of encountered classes through adding samples from the remaining digits in sequence one class at a time. The Calandra network was shown to have overcapacity for just $3$ digits by virtue of its subsequent ability to learn all $10$ digits. We suspect the same overcapacity for Hinton's network and therefore start with a network roughly $1/5$ the size, under the assumption that neurogenesis will grow a network sufficient to learn the remaining digits as they are individually presented for training. Thus the size of our initial DNN prior to neurogenesis was: $784$-$200$-$100$-$75$-$20$-$75$-$100$-$200$-$784$. 
Accordingly, we trained a $1$,$7$-network using all training samples of $1$'s and $7$'s with a stacked denoising AE. After training the $1$,$7$-AE, it is ready to address drifting inputs through NDL. New classes of digits are presented in the following order: $0$, $2$, $3$, $4$, $5$, $6$, $8$, and $9$. Notably, this procedure is not strictly concept drift (where classes are changing over time) or transfer learning (where a trained network is retrained to apply to a different domain), but rather was designed to examine the capability of the network to learn novel inputs while maintaining the stability of previous information (i.e.,~address the stability-plasticity dilemma).

NDL begins by presenting all samples of a new class to Level $1$ of the AE and identifying `outlier' samples having REs above a user-specified threshold. Then, one or more new nodes are added to Level $1$ and the entire level is pre-trained in a SHL-AE. Initially, only the weights connected to the newly added nodes are allowed to be updated at the full learning rate. Encoder weights connected to old nodes are not allowed to change at all (to preserve the feature detectors trained on previous classes) and decoder weights from old nodes are allowed to change at the learning rate divided by $100$. This step relates to the notion of plasticity in biological neurogenesis. After briefly training the new nodes, a stabilization step takes place, where the entire level is trained in a SHL-AE using training samples from all classes seen by the network (samples from old classes are generated via intrinsic replay). After again calculating the RE on samples from the new class, additional nodes are added until either 1) the RE for enough samples falls below the threshold or 2) a user-specified maximum number of new nodes are reached for the current level. Once neurogenesis is complete for a level, weights connecting to the next level are trained using samples from all classes. This process repeats for each succeeding level of the AE using outputs from the previous encoding layer. After NDL, the new AE should be capable of reconstructing images from the new class (e.g.,~$0$) in addition to the current previous classes (e.g.,~$1$ and $7$).

\subsection{Neurogenesis allow encoding of novel information}
\begin{figure}[h]
\begin{center}
\includegraphics[width=3.3in]{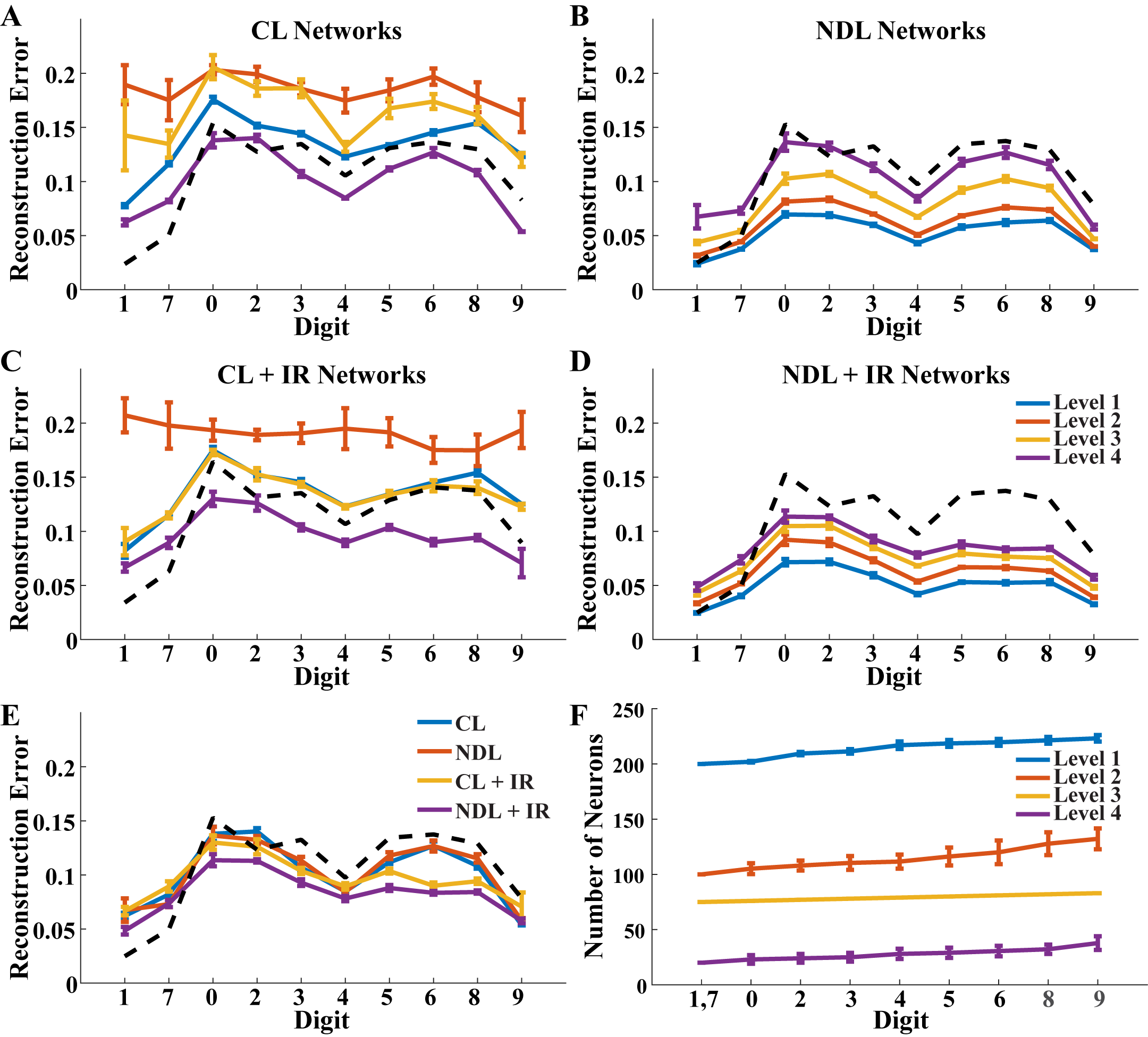}
\caption{Global reconstructions of trained MNIST digits after exposure to all $10$ digits. The legend in Plot D applies to Plots A, B, and C; the dotted line shows REs of the original AE trained just on $1$ and $7$. (A) CL without IR provides only marginal improvement in reconstruction ability after learning all 10 digits; (B) NDL without IR likewise fails to improve reconstruction, though NDL training makes reconstruction through partial networks more useful; (C) CL with IR improves overall reconstruction of previously trained digits; (D) NDL with IR further improves on CL with IR in (C) along with improved partial network reconstructions; (E) Full network reconstructions of all networks after progressive training through all digits; (F) Neurogenesis contribution to network size in NDL+IR networks. \label{comboFigure}}
\end{center}
\end{figure}

Results of NDL experiments on MNIST data showed that an established network trained on just digits $1$ and $7$ can be enlarged through neurogenesis to represent new digits as guided by RE at each level of a stacked AE. We compared a network created with NDL and IR (`NDL+IR') to three control networks: Control $1$ (`CL') – an AE the same size as the enlarged NDL without IR network trained first on the subset digits $1$ and $7$, and then retrained without intrinsic replay on all samples from one new single digit at a time (Figure~\ref{comboFigure}A); Control $2$ (`NDL') – continuous learning on the original $1$,$7$ network using NDL, but not using intrinsic replay (Figure~\ref{comboFigure}B), and Control $3$ (`CL+IR') – an AE the same size as the enlarged NDL+IR network trained first on the subset digits $1$ and $7$, and then retrained with all samples from one new single digit at a time, while using intrinsic replay to generate samples of previously trained classes throughout the experiment (Figure~\ref{comboFigure}C). Figure~\ref{comboFigure}D shows that the network built upon NDL+IR slightly outperforms learning on a fixed network (Figure~\ref{comboFigure}C). Notably, NDL+IR outperforms straight learning not only on reconstruction across all digits, but in both the ability to represent the new data as well as preserving the ability to represent previously trained digits (Figure~\ref{comboFigure}E). This latter point is important, because while getting a trained network to learn new information is not particularly challenging, getting it to preserve old information can be quite difficult.

Note that the final DNN size is unknown prior to neurogenesis. The network size is increased based on the RE when the network is exposed to new information, so there is possible value in using this method to determine an effective DNN size. The original size of the $1$,$7$-AE is $784$-$200$-$100$-$75$-$20$-$75$-$100$-$200$-$784$. Figure~\ref{comboFigure}F shows how the DNN grows as new classes are presented during neurogenesis, gaining more representational capacity as new classes are learned. 

\begin{figure}
\begin{center}
\includegraphics[width=3.3in]{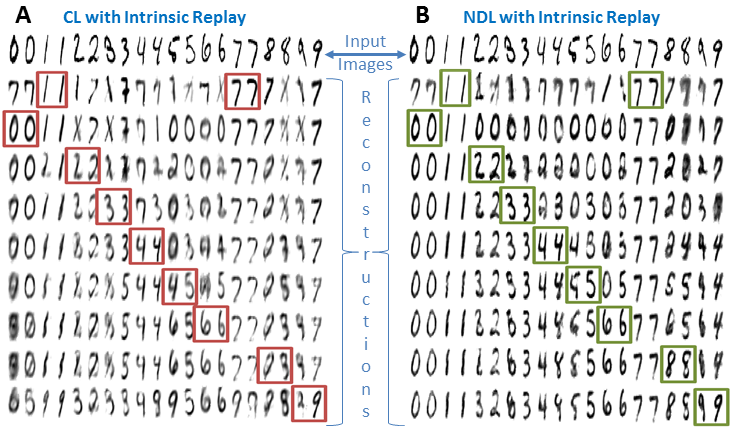}
\caption{Reconstructions of all digits by pre-trained `$1$/$7$' networks after learning on progressive new classes. (A) Networks using conventional learning with IR are able to acquire new digits and show some ability to maintain representations of recently trained digits (e.g.,~'$6$'s after `$8$' is learned). (B) Networks using NDL with IR are able to acquire new digits and show superior reconstructions of previously encountered digits, even for those digits trained far earlier (e.g.,~'$0$'s throughout the experiment). \label{reconstructionFigure}}
\end{center}
\end{figure}
\begin{figure*}[t]
\begin{center}
\includegraphics[width=6.5in]{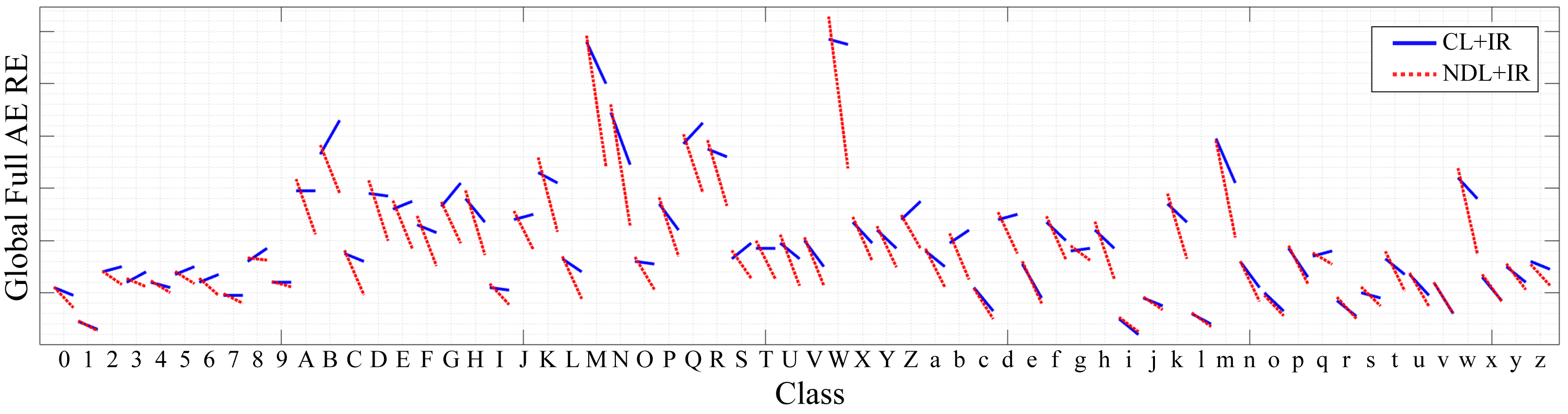}
\caption{Comparison of initial and final (left and right of each line segment, respectively) full AE (Level $4$) REs on each class after learning of all classes.\label{nistFigure}}
\end{center}
\end{figure*}
\begin{figure*}[t]
\begin{center}
\includegraphics[width=6.5in]{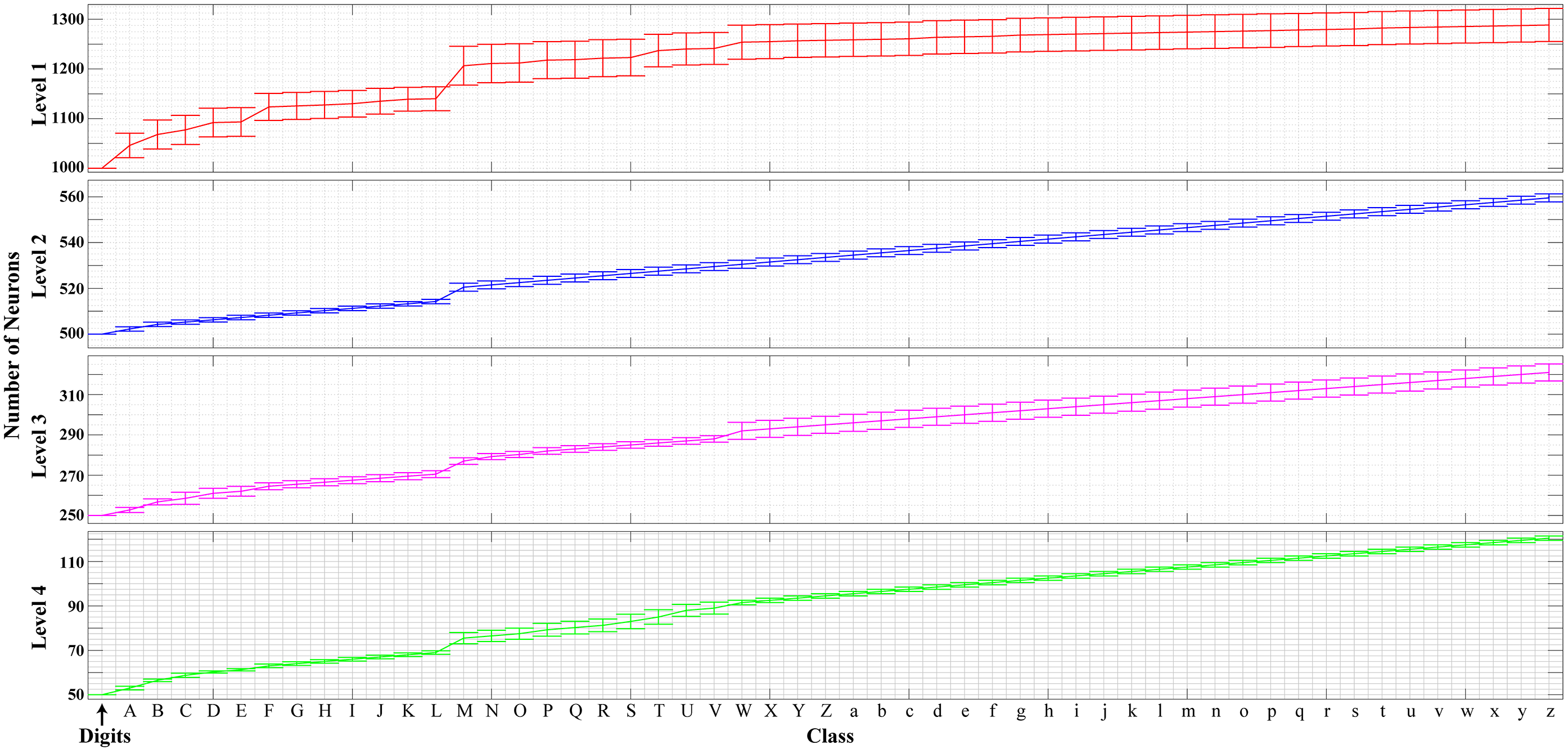}
\caption{Neurogenesis contribution to networks trained on digits only, where new classes are presented for NDL alphabetically, upper case first.\label{ngcontributionFigure}}
\end{center}
\end{figure*}
The `CL+IR' control network initially had the identical size of the neurogenesis network `NDL+IR', was initially trained on digits $1$ and $7$, and then learned to represent the remaining MNIST digits, one at a time in the same order as presented during neurogenesis, but the network size was fixed. Figure~\ref{reconstructionFigure}A shows reconstructed images after each new class was learned on the `CL+IR' AE and Figure~\ref{reconstructionFigure}B shows the comparable images for the `NDL+IR' network as it was trained to accommodate all MNIST digits. One can see that before being trained on new digits (to the right of the blocked trained class shown in each row), both networks mis-reconstructed digits from the unseen classes into digits that appear to belong to a previously trained class as expected. Notably in the `CL+IR' reconstructions (Figure~\ref{reconstructionFigure}A), digits from previously seen classes were often mis-reconstructed to more recently seen classes. In contrast, the `NDL+IR' networks (Figure~\ref{reconstructionFigure}B) were more stable in their representations of previously encoded data, with only minimal disruption to past classes as new information was acquired. This suggests that adding neurons as a network is exposed to novel information is advantageous in maintaining the stability of a DNN's previous representations.

\section{Results on NIST SD $19$}
Applying NDL to the NIST SD $19$ dataset presents challenges for evaluating neurogenesis performance because of the number of classes. Figure~\ref{nistFigure} shows the effect of learning on each class, comparing the initial RE of each class on the network trained on digits before any learning of letters and the final RE after all classes have been learned. A line segment with a downward (negative) slope indicates that the final RE is less than the initial RE. 

The clear observation is that learning new classes with NDL with intrinsic replay (NDL+IR) results in smaller RE than learning without neurogenesis (CL+IR) for all classes. In addition, the final REs for NDL+IR are all lower than the initial REs, even for classes (digits) used to train the original AE. This implies that the ultimate AE built via neurogenesis has a richer set of feature detectors, resulting in better representation of all classes. Another observation is that, in general, the initial REs of the CL+IR network are lower than the initial REs of the NDL+IR network. The reason is that the original NDL+IR network was smaller than the fixed CL+IR network.

\begin{figure*}[t]
\begin{center}
\includegraphics[width=6.5in]{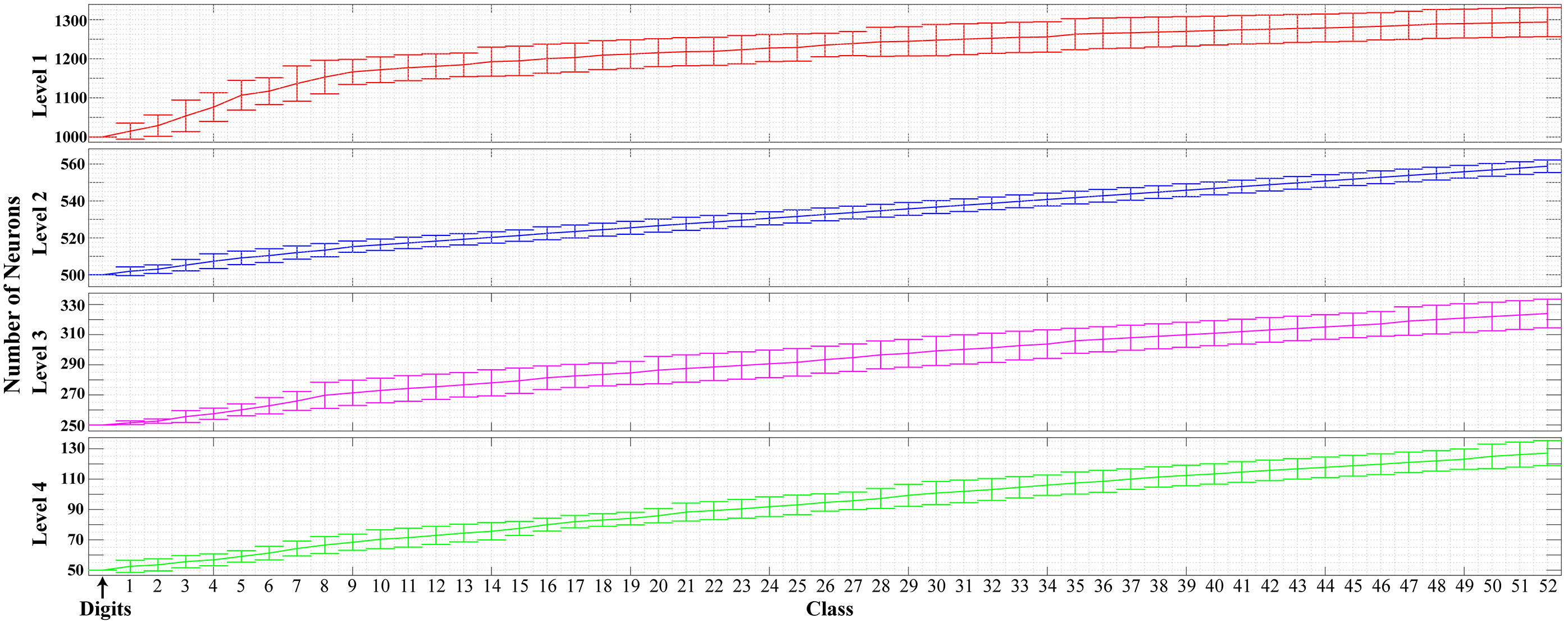}
\caption{Neurogenesis contribution to networks trained on digits only. 20 experiments were conducted where newly presented classes (upper and lower case letters) were randomly ordered. The plot shows the average number of new neurons added progessively for each new class with standard deviation as error bars.\label{ngcontrib2}}
\end{center}
\end{figure*}
While Figure~\ref{nistFigure} shows the improvement in class representation at the beginning and end of NDL+IR, Figures~\ref{ngcontributionFigure} and \ref{ngcontrib2} show the progression in time of growing the final network. More new neurons are added earlier in the neurogenesis process than later. As novel classes are presented, new feature are learned and representation capability improves for all classes. Eventually, the need for additional neurons diminishes. Figure~\ref{ngcontributionFigure} reveals that the AE is particularly lacking feature detectors necessary for good representation of class `M' in all levels. In Figure~\ref{nistFigure}, it is clear that class `W' is also lacking feature detectors, but by the time it is presented for learning, its need has been met by neurogenesis on previous classes.

\section{Characterizing the value of adapting DNNs}
The value of a model to continuously adapt to changing data is challenging to quantify.  Here, we notionally quantify the value of a machine learning algorithm at a given time as 
$U = B - \frac{C_{M}}{\tau} - C_P$,
where the utility, $U$, of an algorithm is considered as a tradeoff between the benefit, $B$, that the computation provides the user, the costs of the algorithm generation or the model itself, $C_M$, and the associated run-time costs, $C_P$, of that computation. $C_P$ typically consists of the time and physical energy and space required for the computation to be performed. For machine learning applications, we must consider the lifetime, $\tau$, of an algorithm for which it is appropriate to amortize a model's build costs. In algorithm design, it is desirable to minimize both of the cost terms; however, the dominant cost will differ depending on the extent to which the real-world data changes. Consider a DNN with $N$ neurons and on the order of $N^2$ synapses. In this example, the cost of building the model, $C_M$, will scale as $O(N^4)$ due to performing $O(N^2)$ operations over $N^2$ training samples during training of a well-regularized, appropriately fit model. As a result, CM will dominate the algorithm's cost unless the lifetime of the model, $\tau$, can offset the polynomial difference between $C_M$ and $C_P$. This description illustrates the need to extend the model's lifetime (e.g.,~via neurogenesis), and to do so in an inexpensive manner that minimizes the data required to adapt the model for future use.

\section{Conclusions and Future Work}
We presented a new adaptive algorithm using the concept of neurogenesis to extend the learning of an established DNN when presented with new data classes. The algorithm adds new neurons at each level of an AE when new data samples are not accurately represented by the network as determined by a high RE. The focus of the paper is on a proof of concept of continuous learning for DNNs to adapt to changing application domains. Several elements of our NDL algorithm that we have not sought to optimize deserve further consideration. For instance, the optimal number of IR samples is unknown and will affect the computational cost associated with their use. Other elements that need to be considered are 1) better ways of establishing and using RE thresholds and 2) developing a method to determine the number of outliers to allow during neurogenesis. While we considered a network of growing size via neurogenesis, adaptation may be obtainable use of a larger network with a fixed size and restricting the learning rate on a subset of neurons until needed at a later time. We evaluated the NDL algorithm on two datasets having gray-scale objects on blank backgrounds and look forward to application on additional datasets, including natural, color imagery.

Ultimately, we anticipate that there are several significant advantages of a neurogenesis-like method for adapting existing networks to incorporate new data, particularly given suitable IR capabilities. The first relates to the costs of DL in application domains. The ability to adapt to new information can extend a model's useful lifetime in real-world situations, possibly by substantial amounts. Extending a model's lifetime increases the duration over which one can amortize the costs of developing the model, and in the case of DL, the build cost often vastly outpaces the runtime operational costs of the trained feed-forward network. As a result, continuous adaptation can potentially make DL cost effective for domains with significant concept drift. Admittedly, the method we describe here has an added processing cost due to the neurogenesis process and the required intrinsic replay; however, this cost will most likely amount to a constant factor increase on the processing costs and still be significantly lower than those costs associated with repeatedly retraining with the original training data. 

The second advantage concerns the continuous learning nature of the NDL algorithm. The ability to train a large network without maintaining a growing repository of data can be valuable, particularly in cases where the bulk storage of data is not permitted due to costs or other restrictions. While much of the DL community has focused on cases where there is extensive unlabeled training data, our technique can provide solutions where training data at any time is limited and new data is expected to arrive continuously. Furthermore, we have considered a very stark change in the data landscape, with the network exposed exclusively to novel classes. In real-world applications, novel information may be encountered more gradually. This slower drift would likely require neurogenesis less often, but it would be equally useful when needed.

Finally, it has not escaped us that the algorithm we present is emulating adult neurogenesis within a cortical-like circuit, whereas in adult mammals, substantial neurogenesis does not appear in sensory cortices~\cite{aimone2014regulation}. In this way, our NDL networks are more similar to juvenile or developmental visual systems, where the network has only been exposed to a limited extent of the information it will eventually encounter. Presumably, if one takes a DNN with many more nodes per layer and trains it with a much larger and broader set of data, the requirement for neurogenesis will diminish. In this situation, we predict that the levels of neurogenesis will eventually diminish to zero early in the network because the DNN will have the ability to represent a broad set of low level features that prove sufficient for even the most novel data encountered, whereas neurogenesis may always remain useful at the deepest network layers that are more comparable to the medial temporal lobe and hippocampus areas of cortex. Indeed, this work illustrates that the incorporation of neural developmental and adult plasticity mechanisms, such as staggering network development by layer (e.g., ``layergenesis''), into conventional DNNs will likely continue to offer considerable benefits.

\subsubsection*{Acknowledgments}
This work was supported by Sandia National Laboratories' Laboratory Directed Research and Development (LDRD) Program under the Hardware Acceleration of Adaptive Neural Algorithms (HAANA) Grand Challenge. Sandia is a multi-mission laboratory managed and operated by Sandia Corporation, a wholly owned subsidiary of Lockheed Martin Corporation, for the U.S. Department of Energy's National Nuclear Security Administration under Contract DE-AC04-94AL85000.

\bibliography{bibliography}

\end{document}